\documentclass[11pt,a4paper]{article}
\PassOptionsToPackage{breaklinks}{hyperref}
\usepackage[hyperref]{acl2019}
\usepackage{times}
\usepackage{latexsym}
\usepackage{booktabs}
\usepackage{url}
\usepackage[flushleft]{threeparttable}

\aclfinalcopy % Uncomment this line for the f inal submission
\title{Racial Bias in Hate Speech and Abusive Language Detection Datasets}

\author{Thomas Davidson \\
  Department of Sociology \\
  Cornell University \\
  \texttt{trd54@cornell.edu} \\\And
  Debasmita Bhattacharya \\
  Department of \\ Computer Science \\
  Cornell University \\
  \texttt{db758@cornell.edu} \\\And
  Ingmar Weber \\
  Qatar Computer \\ Research Institute \\
  \texttt{iweber@hbku.edu.qa}\\}

\date{}

\begin{document}
\maketitle
\begin{abstract}
  %Abusive language detection has become an important area of natural language processing research.
  Technologies for abusive language detection are being developed and applied with little consideration of their potential biases. We examine racial bias in five different sets of Twitter data annotated for hate speech and abusive language. We train classifiers on these datasets and compare the predictions of these classifiers on tweets written in African-American English with those written in Standard American English. The results show evidence of systematic racial bias in all datasets, as classifiers trained on them tend to predict that tweets written in African-American English are abusive at substantially higher rates. If these abusive language detection systems are used in the field they will therefore have a disproportionate negative impact on African-American social media users. Consequently, these systems may discriminate against the groups who are often the targets of the abuse we are trying to detect.
\end{abstract}

\thanks{To appear in the proceedings of the \href{https://sites.google.com/view/alw3/}{Third Abusive Language Workshop} at the Annual Meeting for the Association for Computational Linguistics 2019. \textbf{Please cite the published version.}}

\section{Introduction}

Recent work has shown evidence of substantial bias in machine learning systems, which is typically a result of bias in the training data. This includes both supervised \cite{blodgett_racial_2017, tatman_gender_2017,kiritchenko_examining_2018, de-arteaga_bias_2019} and unsupervised natural language processing systems \cite{bolukbasi_man_2016, caliskan_semantics_2017, garg_word_2018}. Machine learning models are currently being deployed in the field to detect hate speech and abusive language on social media platforms including Facebook, Instagram, and Youtube. The aim of these models is to identify abusive language that directly targets certain individuals or groups, particularly people belonging to protected categories \cite{waseem_understanding_2017}. Bias may reduce the accuracy of these models, and at worst, will mean that the models actively discriminate against the same groups they are designed to protect.

Our study focuses on racial bias in hate speech and abusive language detection datasets \cite{waseem_are_2016, waseem_hateful_2016, davidson_automated_2017, golbeck_large_2017, founta_large_2018}, all of which use data collected from Twitter. We train classifiers using each of the datasets and use a corpus of tweets with demographic information to compare how each classifier performs on tweets written in African-American English (AAE) versus Standard American English (SAE) \cite{blodgett_demographic_2016}. We use bootstrap sampling \cite{efron_bootstrap_1986} to estimate the proportion of tweets in each group that each classifier assigns to each class. We find evidence of systematic racial biases across all of the classifiers, with AAE tweets predicted as belonging to negative classes like hate speech or harassment significantly more frequently than SAE tweets. In most cases the bias decreases in magnitude when we condition on particular keywords which may indicate membership in negative classes, yet it still persists. We expect that these biases will result in racial discrimination if classifiers trained on any of these datasets are deployed in the field. %After describing our results we discuss our findings and how they vary for each of the datasets evaluated.  We conclude by discussing the implications of our study, along with its limitations.

\section{Related works}

Scholars and practitioners have recently been devoting more attention to bias in machine learning models, particularly as these models are becoming involved in more and more consequential decisions \cite{athey_beyond_2017}. Bias often derives from the data used to train these models. For example, \citet{buolamwini_gender_2018} show how facial recognition technologies perform worse for darker-skinned people, particularly darker-skinned women, due to the disproportionate presence of white, male faces in the training data. Natural language processing systems also inherit biases from the data they were trained on. For example, in unsupervised learning, word embeddings often contain biases \cite{bolukbasi_man_2016,caliskan_semantics_2017,garg_word_2018} which persist even after attempts to remove them \cite{gonen_lipstick_2019}. There are many examples of bias in supervised learning contexts: YouTube's captioning models make more errors when transcribing women \cite{tatman_gender_2017}, AAE is more likely to be misclassified as non-English by widely used language classifiers \cite{blodgett_racial_2017}, numerous gender and racial biases exist in sentiment classification systems \cite{kiritchenko_examining_2018}, and errors in both co-reference resolution systems and occupational classification models reflect gendered occupational patterns \cite{zhao_gender_2018, de-arteaga_bias_2019}. 

 While hate speech and abusive language detection has become an important area for natural language processing research \cite{schmidt_survey_2017, waseem_understanding_2017, fortuna_survey_2018}, there has been little work addressing the potential for these systems to be biased. The danger posed by bias in such systems is, however, particularly acute, since it could result in negative impacts on the same populations the systems are designed to protect. For example, if we mistakenly consider speech by a targeted minority group as abusive we might unfairly penalize the victim, but if we fail to identify abuse against them we will be unable to take action against the perpetrator. Although no model can perfectly avoid such problems, we should be particularly concerned about the potential for such models to be \textit{systematically} biased against certain social groups, particularly protected classes. 

A number of studies have shown that false positive cases of hate speech are associated with the presence of terms related to race, gender, and sexuality \cite{kwok_locate_2013, burnap_cyber_2015, davidson_automated_2017}. While not directly measuring bias, prior work has explored how annotation schemes \cite{davidson_automated_2017} and the identity of the annotators \cite{waseem_are_2016} might be manipulated to help to avoid bias. \citet{dixon_measuring_2018} directly measured biases in the Google Perspective API classifier,\footnote{\url{https://www.perspectiveapi.com}} trained on data from Wikipedia talk comments, finding that it tended to give high toxicity scores to innocuous statements like ``I am a gay man''. They called this ``false positive bias'', caused by the model overgeneralizing from the training data, in this case from examples where ``gay'' was used pejoratively. They find that a number of such ``identity terms'' are disproportionately represented in the examples labeled as toxic. \citet{park_reducing_2018} build upon this study, using templates to study gender differences in performance across two hate speech and abusive language detection datasets. They find that classifiers trained on these data tend to perform worse when female identity terms used, indicating gender bias in performance. We build upon this work by auditing a series of abusive language and hate speech detection datasets for racial biases. We evaluate how classification models trained on these datasets perform in the field, comparing their predictions for tweets written in language used by whites or African-Americans.

\section{Research design}

 %In the following section we describe the different datasets to be evaluate, how we trained classifiers with the datasets, the dataset we use to measure racial bias, and finally the three experiments we conducted.

\subsection{Hate speech and abusive language datasets}

We focus on Twitter, the most widely used data source in abusive language research. We use all available datasets where tweets are labeled as various types of abuse and are written in English. We now briefly describe each of these datasets in chronological order.

\citet{waseem_hateful_2016} collected 130k tweets containing one of seventeen different terms or phrases they considered to be hateful. They then annotated a sample of these tweets themselves, using guidelines inspired by critical race theory. These annotators were then reviewed by ``a 25 year old woman studying gender studies and a nonactivist feminist'' to check for bias. This dataset consists of 16,849 tweets labeled as either racism, sexism, or neither. Most of the tweets categorized as sexist relate to debates over an Australian TV show and most of those considered as racist are anti-Muslim. 

To account for potential bias in the previous dataset, \citet{waseem_are_2016} relabeled 2876 tweets in the dataset, along with a new sample from the tweets originally collected. The tweets were annotated by ``feminist and anti-racism activists'', based upon the assumption that they are domain-experts. A fourth category, racism \textit{and} sexism was also added to account for the presence of tweets which exhibit both types of abuse. The dataset contains 6,909 tweets.

\citet{davidson_automated_2017} collected tweets containing terms from the Hatebase,\footnote{\url{https://hatebase.org/}} a crowdsourced hate speech lexicon, then had a sample coded by crowdworkers located in the United States. To avoid false positives that occurred in prior work which considered all uses of particular terms as hate speech, crowdworkers were instructed not to make their decisions based upon any words or phrases in particular, no matter how offensive, but on the overall tweet and the inferred context. The dataset consists of 24,783 tweets annotated as hate speech, offensive language, or neither.

\citet{golbeck_large_2017} selected tweets using ten keywords and phrases related to anti-black racism, Islamophobia, homophobia, anti-semitism, and sexism. The authors developed a coding scheme to distinguish between potentially offensive content and serious harassment, such as threats or hate speech. After an initial round of coding, where tweets were assigned to a number of different categories, they simplified their analysis to include a binary harassment or non-harassment label for each tweet. The dataset consists of 20,360 tweets, each hand-labeled by the authors.\footnote{The paper describes 35k tweets but there were many duplicates in this dataset which were removed from the dataset the authors made available.}

\citet{founta_large_2018} constructed a dataset intended to better approximate a real-world setting where abuse is relatively rare. They began with a random sample of tweets then augmented it by adding tweets containing one or more terms from the Hatebase lexicon and that had negative sentiment. They criticized prior work for defining labels in an ad hoc manner. To develop a more comprehensive annotation scheme they initially labeled a sample of tweets, allowing each tweet to belong to multiple classes. After analyzing the overlap between different classes they settled on a coding scheme with four distinct classes: abusive, hateful, spam, and normal. We use a dataset they published containing 91,951 tweets coded into these categories by crowdworkers.\footnote{They describe 80k tweets in the paper but more tweets were added to the dataset released by the authors. Some of the tweets in this dataset are duplicates: if all versions of a duplicated tweet were coded in the same way by the majority of coders we retained one copy and deleted the rest; if the labels disagreed we removed all copies.}

%Additionally, we also use the Perspective API, developed by Jigsaw and Google. Unlike the other datasets the data are not from Twitter but come from comments tagged by humans as being toxic, and in some cases, severely toxic. The API provides access to a deep convolutional neural network trained using word-vector representations of comments. Unlike the other datasets, for which we trained classifiers, we directly use the output of this pre-trained model in our analysis. %https://github.com/conversationai/perspectiveapi/blob/master/api_reference.md

\begin{table}[ht]
\small
\begin{tabular}{lllll}
Dataset         & Class             & Precision & Recall & F1   \\\hline
\textit{W. \& H.}  & Racism            & 0.73      & 0.79   & 0.76 \\
                & Sexism            & 0.69      & 0.73   & 0.71 \\
                & Neither           & 0.88      & 0.85   & 0.86 \\
\textit{W.}          & Racism            & 0.56      & 0.77   & 0.65 \\
                & Sexism            & 0.62      & 0.73   & 0.67 \\
                & R. \& S. & 0.56      & 0.62   & 0.59 \\
                & Neither           & 0.95      & 0.92   & 0.94 \\
\textit{D. et al.} & Hate       & 0.32      & 0.53   & 0.4  \\
                & Offensive         & 0.96      & 0.88   & 0.92 \\
                & Neither           & 0.81      & 0.95   & 0.87 \\
\textit{G. et al.}  & Harass.        & 0.41      & 0.19   & 0.26 \\
                & Non.     & 0.75      & 0.9    & 0.82 \\
\textit{F. et al.}   & Hate       & 0.33      & 0.42   & 0.37 \\
                & Abusive            & 0.87      & 0.88   & 0.88 \\
                & Spam              & 0.5       & 0.7    & 0.58 \\
                & Neither           & 0.88      & 0.77   & 0.82
\end{tabular}
\caption{\textbf{Classifier performance}}\label{tab:classifier_performance}
\end{table}

\subsection{Training classifiers}

For each dataset we train a classifier to predict the class of unseen tweets. We use regularized logistic regression with bag-of-words features, a commonly used approach in the field. While we expect that we could improve predictive performance by using more sophisticated classifiers, we expect that any bias is likely a function of the training data itself rather than the classifier. %, so we decided this was unnecessary.
Moreover, although features like word embeddings can work well for this task \cite{djuric_hate_2015} we wanted to avoid inducing any bias in our models by using pre-trained embeddings \cite{park_reducing_2018}.

We pre-process each tweet by removing excess white-space and replacing URLs and mentions with placeholders. We then tokenize them, stem each token, and construct n-grams with a maximum length of three. Next we transform each dataset into a TF-IDF matrix, with a maximum of 10,000 features. We use 80\% of each dataset to train models and hold out the remainder for validation. Each model is trained using stratified 5-fold cross-validation. We conduct a grid-search over different regularization strength parameters to identify the best performing model. Finally, for each dataset we identify the model with the best average F1 score and retrain it using all of the training data. The performance of these models on the 20\% held-out validation data is reported in Table~\ref{tab:classifier_performance}. Overall we see varying performance across the classifiers, with some performing much better out-of-sample than others. In particular, we see that hate speech and harassment are particularly difficult to detect. Since we are primarily interested in \textit{within classifier, between corpora} performance, any variation between classifiers should not impact our results.

%TODO: Need to actually make Table 1
%TODO: Need to add information on the Perspective API

%For the Perspective classifier we simply make an API call using the text of the tweet.

%TODO: how did we determine the best model? 
\subsection{Race dataset}
We use a dataset of tweets labeled by race from \citet{blodgett_demographic_2016} to measure racial biases in these classifiers. They collected geo-located tweets in the U.S. and matched them with demographic data from the Census on the population of non-Hispanic whites, non-Hispanic blacks, Hispanics, and Asians in the block group where the tweets originated.
They then identified words associated with particular demographics and trained a probabilistic mixed-membership language model. This model learns demographically-aligned language models for each of the four demographic categories and is used to calculate the posterior proportion of language from each category in each tweet. Their validation analyses indicate that tweets with a high posterior proportion of non-Hispanic black language exhibit lexical, phonological, and syntactic variation consistent with prior research on AAE. Their publicly-available dataset contains 59.2 million tweets.

We define a \textit{user} as likely non-Hispanic black if the average posterior proportion across all of their tweets for the non-Hispanic black language model is $\geq 0.80$ (and $\leq 0.10$ Hispanic and Asian combined) and as non-Hispanic white using the same formula but for the white language model.\footnote{We use this threshold following \citet{blodgett_racial_2017} and after consulting with the lead author. While these cut-offs should provide high confidence that the users tend to use AAE or SAE, and hence serve as a proxy for race, it is important to note that not all African-Americans use AAE and that not all AAE users are African-American, although use of the AAE dialect suggests a social proximity to or affinity for African-American communities \cite{blodgett_demographic_2016}} This allows us to restrict our analysis to tweets written by users who predominantly use one of the language models. Due to space constraints we discard users who predominantly use either the Hispanic or the Asian language model. This results in a set of 1.1m tweets written by people who generally use non-Hispanic black language and 14.5m tweets written by users who tend to use non-Hispanic white language. Following \citet{blodgett_racial_2017}, we call these datasets \textit{black-aligned} and \textit{white-aligned} tweets, reflecting the fact that they contain language associated with either demographic category but which may not all be produced by members of these categories. We now describe how we use these data in our experiments.

\begin{table*}[ht]
\small
\begin{center}
\begin{tabular}{llrrrlr}
\toprule
         Dataset &              Class &  $\widehat{p_{i_{black}}}$ &  $\widehat{p_{i_{white}}}$ &        $t$ &    $p$ &  $\frac{\widehat{p_{i_{black}}}}{\widehat{p_{i_{white}}}}$ \\\hline
                 &                    &        &        &          &      &        \\
\midrule
 \textit{Waseem and Hovy} &             Racism &  0.001 &  0.003 &  -20.818 &  *** &  0.505 \\
 &             Sexism &  0.083 &  0.048 &  101.636 &  *** &  1.724 \\
 %waseem\_and\_hovy &            neither &  0.916 &  0.949 &  -95.573 &  *** &  0.965 \\
          \textit{Waseem} &             Racism &  0.001 &  0.001 &    0.035 &      &  1.001 \\
           &             Sexism &  0.023 &  0.012 &   64.418 &  *** &  1.993 \\
           &  Racism and sexism &  0.002 &  0.001 &    4.047 &  *** &  1.120 \\
          %waseem &            neither &  0.975 &  0.987 &  -63.315 &  *** &  0.988 \\
  \textit{Davidson et al.} &               Hate &  0.049 &  0.019 &  120.986 &  *** &  2.573 \\
   &          Offensive &  0.173 &  0.065 &  243.285 &  *** &  2.653 \\
  %davidson\_et\_al &            neither &  0.778 &  0.916 & -275.346 &  *** &  0.850 \\
   %golbeck\_et\_al &      nonharassment &  0.968 &  0.977 &  -39.483 &  *** &  0.991 \\
   \textit{Golbeck et al.} &         Harassment &  0.032 &  0.023 &   39.483 &  *** &  1.396 \\
    \textit{Founta et al.} &               Hate &  0.111 &  0.061 &  122.707 &  *** &  1.812 \\
     &            Abusive &  0.178 &  0.080 &  211.319 &  *** &  2.239 \\
    %founta\_et\_al &            neither &  0.682 &  0.844 & -266.970 &  *** &  0.809 \\
     &               Spam &  0.028 &  0.015 &   63.131 &  *** &  1.854 \\
\bottomrule
\end{tabular}
\caption{\textbf{Experiment 1}}\label{tab:experiment1}
\begin{tablenotes}
      \small
      \item We focus on the ``negative'' classes so other classes have been omitted. Stars indicate level of statistical significance. $*** = p < 0.001.$ No stars indicates $p > 0.05.$ 
    \end{tablenotes}
\end{center}
\end{table*}

\subsection{Experiments}
We examine whether the probability that a tweet is predicted to belong to a particular class varies in relation to the racial alignment of the language it uses.
The null hypothesis of no racial bias is that the probability a tweet will belong to a negative class is independent of the racial group the tweet's author is a member of. Formally, for class $c_{i}$, where $c_{i} = 1$ denotes membership in the class and $c_{i} = 0$ the opposite, we aim to test $H_{N} : P(c_{i}=1 | black) = P(c_{i}=1 | white)$. If  $P(c_{i}=1 | black) > P(c_{i}=1 | white)$ and the difference is statistically significant then we can reject the null hypothesis $H_{N}$ in favor of the alternative hypothesis $H_{A}$ that black-aligned tweets are classified into $c_{i}$ at a higher rate than white-aligned tweets. Conversely, if  $P(c_{i}=1 | black) < P(c_{i}=1 | white)$ we can conclude that the classifier is more likely to classify white-aligned tweets as $c_{i}$.  We should expect that white-aligned tweets are more likely to use racist language or hate speech than black-aligned tweets, given that African-Americans are often targeted with racism and hate speech by whites. However for some classes like sexism we have no reason to expect there to be racial differences in either direction. %The ratio $\frac{P(c_{i}=1 | black)}{P(c_{i}=1 | white)}$ provides a measure of the rate at which tweets in the black-aligned corpus are classified into category $c_{i}$ compared to whites.

To test this hypothesis we use bootstrap sampling \cite{efron_bootstrap_1986} to estimate the proportion of tweets in each dataset that each classifier predicts to belong to each class. We draw $n$ random samples with replacement of $k$ tweets from each of the two race corpora, where $n = k = 1000$. For each sample we use each classifier to predict the class membership of each tweet, then store the proportion of tweets that were assigned to each class, $p_{i}$. For each classifier-class pair, we thus obtain a pair of vectors, one for each corpus, each containing $n$ sampled proportions. The bootstrap estimates for the proportion of tweets belonging to class $i$ for each group, $\widehat{p_{i_{black}}}$ and $\widehat{p_{i_{white}}}$, are calculated by taking the mean of the elements in each vector: $\frac{1}{n} \sum_{j=1}^{n} p_{ij}$. We then use a t-test to test whether $\widehat{p_{i_{black}}} = \widehat{p_{i_{white}}}$. We also calculate the ratio $\frac{\widehat{p_{i_{black}}}}{\widehat{p_{i_{white}}}}$, which shows the magnitude of any difference. Values greater than 1 indicate that black-aligned tweets are classified as belonging to class $i$ at a higher rate than white-aligned tweets. %indicating the presence of bias against African-Americans.

We also conduct a second experiment, where we assess whether there is racial bias conditional upon a tweet containing a keyword likely to be associated with a negative class. While differences in language will undoubtedly remain, this should help to account for the possibility that results in Experiment 1 are driven by differences in the true distribution of the different classes of interest, or of words associated with these classes, in the two corpora. For classifier $c$ and category $i$, we evaluate $H_{N}: P(c_{i}=1 | black, t) = P(c_{i}=1 | white, t)$ for a given term $t$. We conduct this experiment for two different terms, each of which occurs frequently enough in the data to enable our bootstrapping approach. We select the term ``n*gga'', since it is a particularly prevalent source of false positives for hate speech detection \cite{kwok_locate_2013, davidson_automated_2017, waseem_bridging_2018}.\footnote{We also planned to conduct the same analysis using the ``-er'' suffix, however the sample was too small, with the word being used in 555 tweets in the white-aligned corpus (0.004\%) and 61 in the black-aligned corpus (0.005\%).} In this case, we expect that tweets containing the word should be classified as more negative when used by whites, thus $H_{A_{1}}: P(c_{i}=1 | black, t) < P(c_{i}=1 | white, t)$. The other alternative, $H_{A_{2}}: P(c_{i}=1 | black, t) > P(c_{i}=1 | white, t)$ would indicate that black-aligned tweets containing the term are penalized at a higher rate than comparable white-aligned tweets. We also assess the results for the word ``b*tch'' since it is a widely used sexist term, which is often also used casually, but we have no theoretical reason to expect there to be racial differences in its usage. The term ``n*gga'' was used in around 2.25\% of black-aligned and 0.15\% of white-aligned tweets. The term ``b*tch''  was used in 1.7\% of black-aligned and 0.5\% of white-aligned tweets. The substantial differences in the distributions for these two terms alone are consistent with our intuition that some of the results in Experiment 1 may be driven by differences in the frequencies of words associated with negative classes in the training datasets. Since we are using a subsample of the available data, we use smaller bootstrap samples, drawing $k=100$ tweets each time.

\section{Results}
%\subsection{Experiment 1}
The results of Experiment 1 are shown in Table~\ref{tab:experiment1}. We observe substantial racial disparities in the performance of \textit{all classifiers}. In all but one of the comparisons, there are statistically significant ($p < 0.001$) differences %in the estimated proportions at ,
and in all but one of these we see that tweets in the black-aligned corpus are assigned negative labels more frequently than those by whites. The only case where black-aligned tweets are classified into a negative class less frequently than white-aligned tweets is the racism class in the \citet{waseem_hateful_2016} classifier. Note, however, the extremely low rate at which tweets are predicted to belong to this class for both groups. On the other hand, this classifier is 1.7 times more likely to classify tweets in the black-aligned corpus as sexist. For \citet{waseem_are_2016} we see that there is no significant difference in the estimated rates at which tweets are classified as racist across groups, although the rates remain low. Tweets in the black-aligned corpus are classified as containing sexism almost twice as frequently and 1.1 times as frequently classified as containing racism and sexism compared to those in the white-aligned corpus. Moving onto \citet{davidson_automated_2017}, we find large disparities, with around 5\% of tweets in the black-aligned corpus classified as hate speech compared to 2\% of those in the white-aligned set. Similarly, 17\% of black-aligned tweets are predicted to contain offensive language compared to 6.5\% of white-aligned tweets. The classifier trained on the \citet{golbeck_large_2017} dataset predicts black-aligned tweets to be harassment 1.4 times as frequently as white-aligned tweets. The \citet{founta_large_2018} classifier labels around 11\% of tweets in the black-aligned corpus as hate speech and almost 18\% as abusive, compared to 6\% and 8\% of white-aligned tweets respectively. It also classifies black-aligned tweets as spam 1.8 times as frequently.  %TODO: ADD a table containing the performance of the classifiers as Table 1.
%\subsection{Experiment 2}
%\subsubsection{Condition on ``n*gga''}

The results of Experiment 2 are consistent with the previous results, although there are some notable differences. In most cases the racial disparities persist, although they are generally smaller in magnitude and in some cases the direction even changes. Table~\ref{tab:experiment2ngga} shows that for tweets containing the word ``n*gga'', classifiers trained on \citet{waseem_hateful_2016} and \citet{waseem_are_2016} are both predict black-aligned tweets to be instances of sexism approximately 1.5 times as often as white-aligned tweets. The classifier trained on the \citet{davidson_automated_2017}  data is significantly \textit{less} likely to classify black-aligned tweets as hate speech, although it is more likely to classify them as offensive. \citet{golbeck_large_2017} classifies black-aligned tweets as harassment at a higher rate for both groups than in the previous experiment, although the disparity is narrower. For the \citet{founta_large_2018} classifier we see that black-aligned tweets are slightly \textit{less} frequently considered to be hate speech but are much more frequently classified as abusive. %, although for both groups over 90\% of the tweets are being classified into this category.
%vthere are no longer any significant differences with respect to the racism and racism \textit{and} sexism classes, yet in both cases - cut from W&H discussion

\begin{table*}[ht]
\small
\begin{center}
\begin{tabular}{llrrrlr}
\toprule
         Dataset &              Class &  $\widehat{p_{i_{black}}}$ &  $\widehat{p_{i_{white}}}$ &        $t$ &    $p$ &  $\frac{\widehat{p_{i_{black}}}}{\widehat{p_{i_{white}}}}$ \\\hline
                 &                    &        &        &          &      &        \\
\midrule
 \textit{Waseem and Hovy} &             Racism &  0.010 &  0.011 &  -1.462 &      &  0.960 \\
 &             Sexism &  0.147 &  0.100 &  31.932 &  *** &  1.479 \\
% &            neither &  0.851 &  0.898 & -31.302 &  *** &  0.948 \\
          \textit{Waseem} &             Racism &  0.010 &  0.010 &   0.565 &      &  1.027 \\
           &             Sexism &  0.040 &  0.026 &  18.569 &  *** &  1.554 \\
          &  Racism and sexism &  0.011 &  0.010 &   0.835 &      &  1.026 \\
          %waseem &            neither &  0.959 &  0.975 & -20.385 &  *** &  0.983 \\
\textit{Davidson et al.} &               Hate &  0.578 &  0.645 & -31.248 &  *** &  0.896 \\
 &          Offensive &  0.418 &  0.347 &  32.895 &  *** &  1.202 \\
  %&            neither &  0.012 &  0.014 &  -4.881 &  *** &  0.855 \\
   %golbeck\_et\_al &      nonharassment &  0.915 &  0.922 &  -5.984 &  *** &  0.992 \\
   \textit{Golbeck et al.} &         Harassment &  0.085 &  0.078 &   5.984 &  *** &  1.096 \\

    \textit{Founta et al.} &               Hate &  0.912 &  0.930 & -15.037 &  *** &  0.980 \\
     &            Abusive &  0.086 &  0.067 &  16.131 &  *** &  1.296 \\
    %founta\_et\_al &            neither &  0.011 &  0.012 &  -1.543 &      &  0.956 \\
     &               Spam &  0.010 &  0.010 &  -1.593 &      &  1.000 \\
\bottomrule
\end{tabular}
\caption{\textbf{Experiment 2, $t =$ ``n*gga''}}\label{tab:experiment2ngga}
\end{center}
\end{table*}

%\subsubsection{Conditional on ``b*tch''}
The results for the second variation of Experiment 2 where we conditioned on the word ``b*tch'' are shown in Table~\ref{tab:experiment2btch}. We see similar results for \citet{waseem_hateful_2016} and \citet{waseem_are_2016}. In both cases the classifiers trained upon their data are still more likely to flag black-aligned tweets as sexism. The \citet{waseem_hateful_2016} classifier is particularly sensitive to the word ``b*tch'' with 96\% of black-aligned and 94\% of white-aligned tweets predicted to belong to this class. For \citet{davidson_automated_2017} almost all of these tweets are classified as offensive, however those in the black-aligned corpus are 1.15 times as frequently classified as hate speech. We see a very similar result for \citet{golbeck_large_2017} compared to the previous experiment, with black-aligned tweets flagged as harassment at 1.1 times the rate of those in the white-aligned corpus. Finally, for the \citet{founta_large_2018} classifier we see a substantial racial disparity, with black-aligned tweets classified as hate speech at 2.7 times the rate of white aligned ones, a higher rate than in Experiment 1. %For both groups, over 90\% of these tweets were classified as abusive.

\begin{table*}[ht]
\small
\begin{center}
\begin{tabular}{llrrrlr}
\toprule
         Dataset &              Class &  $\widehat{p_{i_{black}}}$ &  $\widehat{p_{i_{white}}}$ &        $t$ &    $p$ &  $\frac{\widehat{p_{i_{black}}}}{\widehat{p_{i_{white}}}}$ \\\hline
                 &                    &        &        &          &      &        \\
\midrule
 \textit{Waseem and Hovy} &             Racism &  0.010 &  0.010 &  -0.632 &      &  0.978 \\
 &             Sexism &  0.963 &  0.944 &  20.064 &  *** &  1.020 \\
 %waseem\_and\_hovy &            neither &  0.038 &  0.055 & -19.092 &  *** &  0.683 \\
    
          \textit{Waseem} &             Racism &  0.011 &  0.011 &  -1.254 &      &  0.955 \\
           &             Sexism &  0.349 &  0.290 &  28.803 &  *** &  1.203 \\
           &  Racism and sexism &  0.012 &  0.012 &  -0.162 &      &  0.995 \\
          %waseem &            neither &  0.646 &  0.705 & -28.597 &  *** &  0.916 \\
  \textit{Davidson et al.} &               Hate &  0.017 &  0.015 &   4.698 &  *** &  1.152 \\
   &          Offensive &  0.988 &  0.991 &  -6.289 &  *** &  0.997 \\
  %davidson\_et\_al &            neither &  0.010 &  0.010 &   1.289 &      &  1.025 \\
  % &      nonharassment &  0.901 &  0.909 &  -6.273 &  *** &  0.991 \\
   \textit{Golbeck et al.} &         Harassment &  0.099 &  0.091 &   6.273 &  *** &  1.091 \\
\textit{Founta et al.} &               Hate &  0.074 &  0.027 &  46.054 &  *** &  2.728 \\
     &            Abusive &  0.925 &  0.968 & -41.396 &  *** &  0.956 \\
    %founta\_et\_al &            neither &  0.010 &  0.014 &  -6.176 &  *** &  0.736 \\
     &               Spam &  0.010 &  0.010 &   0.000 &   &  1.000 \\
\bottomrule
\end{tabular}
\caption{\textbf{Experiment 2, $t =$ ``b*tch'' }}\label{tab:experiment2btch}
\end{center}
\end{table*}

\section{Discussion}
Our results demonstrate consistent, systematic and substantial racial biases in classifiers trained on all five datasets. In almost every case, black-aligned tweets are classified as sexism, hate speech, harassment, and abuse at higher rates than white-aligned tweets. To some extent, the results in the first experiment may be driven by underlying differences in the rates at which speakers of different dialects use particular words and phrases associated with these negative classes in the training data. For example, the word ``n*gga'' appears fifteen times as frequently in the black-aligned corpus compared to the white-aligned corpus.\footnote{It is also possible that these disparities are amplified by the \citet{blodgett_demographic_2016} model, which constructs the posterior proportions of different language models in part by exploiting underlying differences in word frequencies associated with the different demographic categories.} However, the second experiment shows that these disparities tend to persist even when comparing tweets containing keywords likely to be associated with negative classes. While some of the remaining disparities are likely due to differences in the distributions of other keywords we did not condition on, we expect that other more innocuous aspects of black-aligned language may be associated with negative labels in the training data, leading classifiers to disproportionately predict that tweets by African-Americans belong to negative classes. 
We now discuss the results as they pertain to each of the datasets used.

Classifiers trained on data from \citet{waseem_hateful_2016} and \citet{waseem_are_2016} only predicted a small fraction of the tweets to be racism. We suspect that this is due to the composition of their dataset, since the majority of the racist training examples consist of anti-Muslim rather than anti-black language. Across both datasets the words ``n*gger'' and ``n*gga'' appear in 4 and 10 tweets respectively. Looking at the sexism class on the other hand, we see that both models were consistently classifying tweets in the black-aligned corpus as sexism at a substantially higher rate than those in the white-aligned corpus. Given this result, and the gender biases identified in these data by \citet{park_reducing_2018}, it not apparent that the purportedly expert annotators were any less biased than amateur annotators \cite{waseem_are_2016}.

The classifier trained on \citet{davidson_automated_2017} shows the largest disparities in Experiment 1, with tweets in the black-aligned corpus classified as hate speech and offensive language at substantially higher rates than white-aligned tweets. We expect that this result occurred for two reasons. % in particular.
First, the dataset contains a large number of cases where AAE is used \cite{waseem_bridging_2018}. Second, many of the AAE tweets also use words like ``n*gga'' and ``b*tch'', and are thus frequently associated with the hate speech and offensive classes, resulting in ``false positive bias'' \cite{dixon_measuring_2018}. On the other hand, the distinction between hate speech and offensive language appears to hold up to scrutiny: while a large proportion of tweets in Experiment 2 containing the word ``n*gga'' are classified as hate speech, the rate is substantially higher for white-aligned tweets. Without this category we expect that many of the tweets classified as offensive would instead be mistakenly classified as hate speech.

Turning to the \citet{golbeck_large_2017} classifer we found that tweets in the black-aligned dataset were significantly more likely to be classified as harassment in all experiments, although the disparity decreased substantially after conditioning on certain keywords. It seems likely that their simple binary labelling scheme may not be sufficient to capture the variation in language used, resulting in high rates of false positives.

Finally, \citet{founta_large_2018} is the largest and perhaps the most comprehensive of the available datasets. In Experiment 1 we see that this classifier has the second highest rates of racial disparities, classifying black-aligned tweets as hate speech, abusive, and spam at substantially higher rates than white-aligned tweets. In Experiment 2 the classifier is slightly less likely to classify black-aligned tweets containing the word ``n*gga'' as hate speech but is 2.7 times more likely to predict that black-aligned tweets using ``b*tch'' belong to this category.

\section{Conclusion}
Our study is the first to measure racial bias in hate speech and abusive language detection datasets. We find evidence of substantial racial bias in all of the datasets tested. This bias tends to persist even when comparing tweets containing certain relevant keywords. While these datasets are still valuable for academic research, we caution against using them in the field to detect and particularly to take enforcement action against different types of abusive language. If they are used in this way we expect that they will \textit{systematically} penalize African-Americans more than whites, resulting in racial discrimination. We have not evaluated these datasets for bias related to other ethnic and racial groups, nor other protected categories like gender and sexuality, but expect that such bias is also likely to exist.
We recommend that efforts to measure and mitigate bias should start by focusing on how bias enters into datasets as they are collected and labeled. In particular, future work should focus on the following three areas.

First, we expect that some biases emerge at the point of data collection. Some studies sampled tweets using small, ad hoc sets of keywords created by the authors \cite{waseem_hateful_2016, waseem_are_2016, golbeck_large_2017}, an approach demonstrated to produce poor results \cite{king_computer-assisted_2017}. Others start with large crowd-sourced dictionaries of keywords, which tend to include many irrelevant terms, resulting in high rates of false positives \cite{davidson_automated_2017, founta_large_2018}. In both cases, by using keywords to identify relevant tweets we are likely to get non-representative samples of training data that may over- or under-represent certain communities. In particular, we need to consider whether the linguistic markers we use to identify \textit{potentially} abusive language may be associated with language used by members of protected categories. For example, although \citet{davidson_automated_2017} started with thousands of terms from the Hatebase lexicon, AAE is over-represented in the dataset \cite{waseem_bridging_2018} because some keywords associated with this speech community were used more frequently on Twitter than other keywords in the lexicon and were consequentially over-sampled.

Second, we expect that the people who annotate data have their own biases. Since individual biases in reflect societal prejudices, they aggregate into systematic biases in training data. The datasets considered here relied upon a range of different annotators, from the authors \cite{golbeck_large_2017, waseem_hateful_2016} and crowdworkers \cite{davidson_automated_2017, founta_large_2018} to activists \cite{waseem_are_2016}.
Even the classifier trained on expert-labeled data \cite{waseem_are_2016} flags black-aligned tweets as sexist at almost twice the rate of white-aligned tweets. While we agree that there is value in working with domain-experts to annotate data, these results suggest that activists may be prone to similar biases as academics and crowdworkers. Further work is therefore necessary to better understand how to integrate expertise into the process and how training can be used to help to mitigate bias. We also need to consider how sociocultural context influences annotators' decisions. For example, 48\% of the workers employed by \citet{founta_large_2018} were located in Venezuela but the authors did not consider whether this affected their results (or if the annotators understood English sufficiently for the task).

Third, we observed substantial variation in the rates of class membership across classifiers and datasets. In Experiment 1 the rate at which tweets were assigned to negative classes varied from 1\% to
18\%. Some of the low proportions may indicate a preponderance of false negatives due to a lack of training data, suggesting that these models may not be able to sufficiently generalize beyond the data they were trained on. The high proportions may signal too many false positives, which may a result of the over-sampling of abusive language in labeled datasets. \citet{founta_large_2018} claim that, on average, between 0.1\% and 3\% of tweets are abusive, depending upon the category of abuse. Identifying such content is therefore a highly imbalanced classification problem.
When labeling datasets and evaluating our models we must pay more attention to the baseline rates of usage of different types of abusive language and how they may vary across populations \cite{silva_analyzing_2016}. 

Finally, we need to more carefully consider how contextual factors interact with linguistic subtleties and our definitions of abuse. The ``n-word'' is a particularly useful illustration of this issue. It exhibits polysemy, as it can be extremely racist or quotidian, depending on the speaker, the context, and the spelling. While the history of the word and its usages is too complex to be summarized here \cite{neal_nigga:_2013}, when used with the ``-er`` suffix it is generally considered to be a racist ephiphet, associated with white supremacy. Prior work has confirmed that the use of this variant online is generally considered to be hateful \cite{kwok_locate_2013}, although not always the case, for example when a victim of abuse shares an insult they have received. However the variant with the ``-a'' suffix is typically used innocuously by African-Americans \cite{kwok_locate_2013}, indeed our results indicate that it is used far more frequently in black-aligned tweets (although it is still used by many white people).\footnote{This spelling also exhibits derhotacization, a phonological feature of AAE \cite{blodgett_demographic_2016}.} Despite this distinction, some studies have considered this variant to be hateful \cite{silva_analyzing_2016, alorainy_suspended_2018}. This approach results in high rates of false positive cases of hate speech, thus \citet{davidson_automated_2017} included a class for offensive language which does not appear to be hateful and let annotators decide which class tweets belonged to based upon their interpretation of the context, many of whom labeled tweets containing the term as offensive. \citet{waseem_bridging_2018} criticized this decision, claiming that it is problematic to ever consider the word to be offensive due to its widespread use among AAE speakers. This critique appears to be reasonable in the sense that we should not penalize African-Americans for using the word, but it avoids grappling with how to act when the word is used by other speakers and in other contexts. What should be done if it is used by a white social media user in reference to a black user? How should the context of their interaction and the nature of their relationship affect our decision? 

A ``one-size-fits-all'', context-independent approach to defining and detecting abusive language is clearly inappropriate. Different communities have different speech norms, such that a model suitable for one community may discriminate against another. However there is no consensus in the field on how and if we can develop detection systems sensitive to different social and cultural contexts. In addition to our recommendations for improving training data, we emphasize the necessity of considering how context matters and how detection systems will have uneven effects across different communities.

\section{Limitations}
First, while the \citet{blodgett_demographic_2016} dataset is the best available source of tweets labeled as AAE, we do not have ground truth labels for the racial identities of the authors. By filtering on users who predominantly used one type of language we may also miss users who may frequently code-switch between AAE and SAE. Second, although we roughly approximate this in Experiment 2, we cannot rule out the possibility that the results, rather than evidence of bias, are a function of different distributions of negative classes in the corpora studied. It is possible that words associated with negative categories in our abusive language datasets are also used to predict race by \citet{blodgett_demographic_2016}, potentially contributing to the observed disparities. To more thoroughly investigate this issue we therefore require ground truth labels for abuse \textit{and} race. Third, the results may vary for different classifiers or feature sets. It is possible that more sophisticated modeling approaches could enable us to alleviate bias, although they could also exacerbate it. Fourth, we did not interpret the results of the classifiers to determine why they made particular predictions. Further work is needed to identify
what features of AAE the classifiers are learning to associate with negative classes.  Finally, this study has only focused on one dimension of racial bias. Further work is necessary to assess the degree to investigate the extent to which data and models are biased against people belonging to other protected categories.

\section*{Acknowledgments}
We would like to thank Jonathan Chang, Emily Parker, Ben Rosche, and the four anonymous reviewers for their comments and suggestions. We also thank the authors of the datasets used for making their data available, particularly Su Lin Blodgett, Antigoni Founta, Jennifer Golbeck, and Zeerak Waseem, who diligently responded to our queries.

\bibliography{acl2019}
\bibliographystyle{acl_natbib}

\appendix

\end{document}